\documentclass[runningheads]{llncs}
\usepackage[T1]{fontenc}
\usepackage{graphicx}
\usepackage{amsmath}
\usepackage{cite}
\usepackage{amssymb}

\usepackage{booktabs}
\usepackage{adjustbox}
\usepackage[table,xcdraw]{xcolor}
\usepackage{subcaption}
\usepackage[unicode=true, bookmarks=true, breaklinks=false, pdfborder={0 0 1}, backref=false, colorlinks=false]{hyperref}
\newcommand{\orcid}[1]{\href{https://orcid.org/#1}{\includegraphics[scale=0.02]{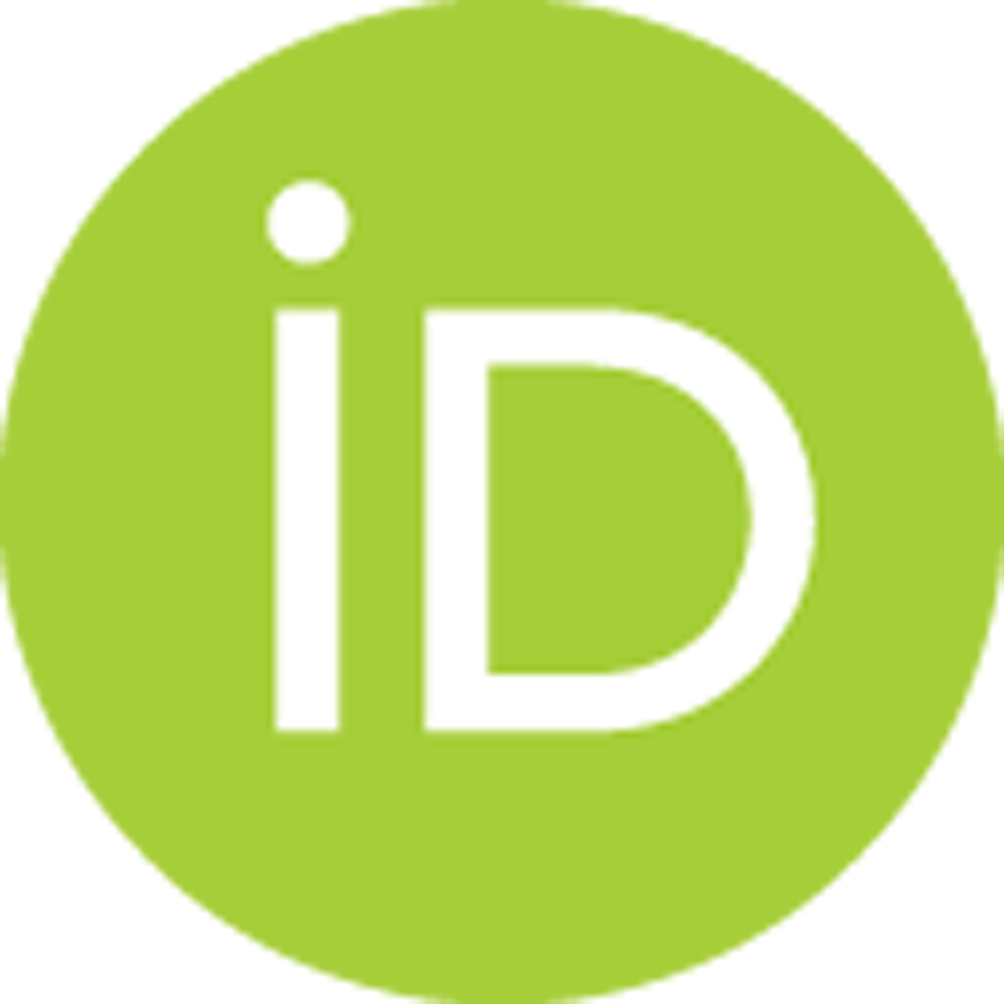}}}

\begin{document}
\title{Diffusion-Based Hierarchical Multi-Label Object Detection to Analyze Panoramic Dental X-rays}

\titlerunning{Diffusion-Based Hierarchical Multi-Label Detection}
% If the paper title is too long for the running head, you can set
% an abbreviated paper title here

    \author{ 
    Ibrahim Ethem Hamamci\inst{1*}
    \and
    Sezgin Er\inst{2}
    \and
    Enis Simsar\inst{3}
    \and
    Anjany Sekuboyina\inst{1}
    \and
    Mustafa Gundogar\inst{4}
    \and
    Bernd Stadlinger\inst{5}
    \and
    Albert Mehl\inst{5}
    \and
    Bjoern Menze\inst{1}}

    %\author{ 
    %Ibrahim Ethem Hamamci\inst{1,2*\orcid{0000-0003-2932-3105}}
   % \and
   % Sezgin Er\inst{2\orcid{0000-0001-7266-9844}}
    %\and
    %Enis Simsar\inst{3\orcid{0000-0002-6662-3249}}
    %\and
    %Anjany Sekuboyina\inst{1\orcid{0000-0002-5601-284X}}
    %\and
    %Mustafa Gundogar\inst{4}
    %\and
    %Bernd Stadlinger\inst{5\orcid{0000-0001-5044-7052}}
    %\and
    %Albert Mehl\inst{5}
    %\and
    %Bjoern Menze\inst{1\orcid{0000-0003-4136-5690}}}

    \institute{Department of Quantitative Biomedicine, University of Zurich, Switzerland
    \and
    International School of Medicine, Istanbul Medipol University, Turkey
    \and
    Department of Computer Science, ETH Zurich, Switzerland
    \and
    Department of Endodontics, Istanbul Medipol University, Turkey
    \and
    Center of Dental Medicine, University of Zurich, Switzerland
    \\
    ~ 
    \\
    \textbf{*} Corresponding author\\
    \email{\{ibrahim.hamamci@uzh.ch\}}}

\authorrunning{Hamamci et al.}

\maketitle              % typeset the header of the contribution
\begin{abstract}
    Due to the necessity for precise treatment planning, the use of panoramic X-rays to identify different dental diseases has tremendously increased. Although numerous ML models have been developed for the interpretation of panoramic X-rays, there has not been an end-to-end model developed that can identify problematic teeth with dental enumeration and associated diagnoses at the same time. To develop such a model, we structure the three distinct types of annotated data hierarchically following the FDI system, the first labeled with only quadrant, the second labeled with quadrant-enumeration, and the third fully labeled with quadrant-enumeration-diagnosis. To learn from all three hierarchies jointly, we introduce a novel diffusion-based hierarchical multi-label object detection framework by adapting a diffusion-based method that formulates object detection as a denoising diffusion process from noisy boxes to object boxes. Specifically, to take advantage of the hierarchically annotated data, our method utilizes a novel noisy box manipulation technique by adapting the denoising process in the diffusion network with the inference from the previously trained model in hierarchical order. We also utilize a multi-label object detection method to learn efficiently from partial annotations and to give all the needed information about each abnormal tooth for treatment planning. Experimental results show that our method significantly outperforms state-of-the-art object detection methods, including RetinaNet, Faster R-CNN, DETR, and DiffusionDet for the analysis of panoramic X-rays, demonstrating the great potential of our method for hierarchically and partially annotated datasets. The code and the datasets are available at \url{https://github.com/ibrahimethemhamamci/HierarchicalDet}.
\end{abstract}
\keywords{Diffusion Network, Hierarchical Learning, Multi-Label Object Detection, Panoramic Dental X-ray, Transformers}

\section{Introduction}
\label{section:introduction}
The use of panoramic X-rays to diagnose numerous dental diseases has increased exponentially due to the demand for precise treatment planning~\cite{Hwang2019}. However, visual interpretation of panoramic X-rays may consume a significant amount of essential clinical time~\cite{Bruno2015} and interpreters may not always have dedicated training in reading scans as specialized radiologists have~\cite{Kumar2021}. Thus, the diagnostic process can be automatized and enhanced by getting the help of Machine Learning~(ML) models. For instance, an ML model that automatically detects abnormal teeth with dental enumeration and associated diagnoses would provide a tremendous advantage for dentists in making decisions quickly and saving their time.
\begin{figure}[h!]
    \vspace{-0.2cm}
    \includegraphics[width=1\linewidth]{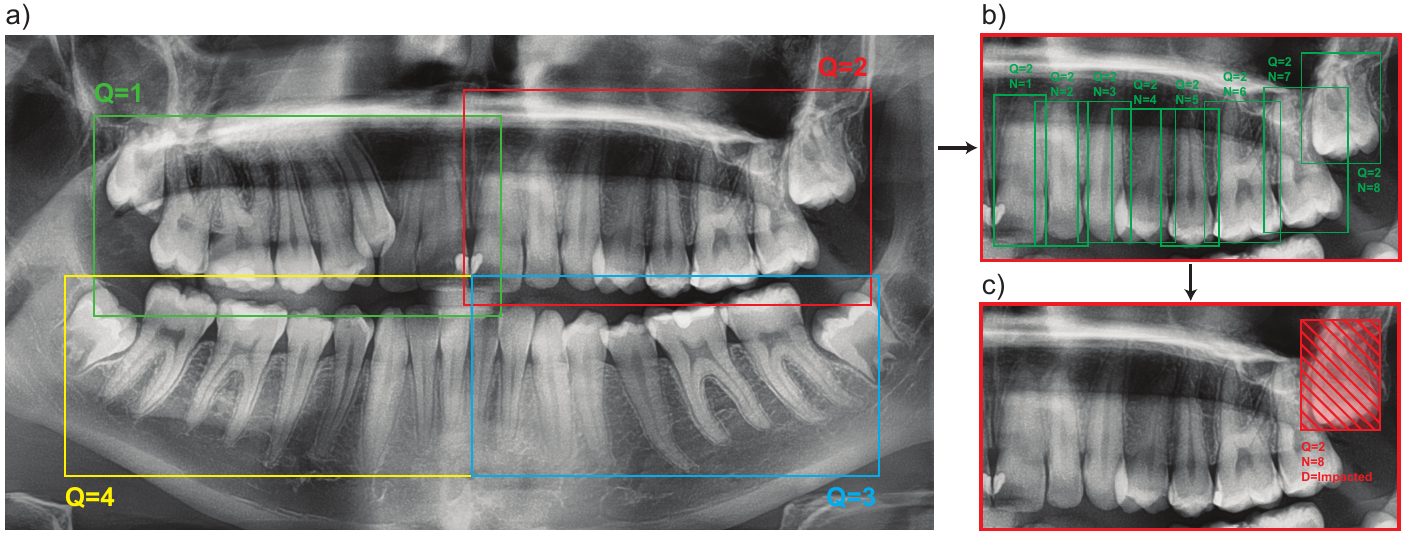}
    \centering
    \caption{The annotated datasets are organized hierarchically as (a) quadrant-only, (b) quadrant-enumeration, and (c) quadrant-enumeration-diagnosis respectively.} 
    %following the FDI system, featuring full labels including quadrant, enumeration, and diagnosis, as well as partial labels with quadrant-only and quadrant-enumeration information.} 
    \label{fig2}
    \vspace{-0.1cm}
\end{figure}

Many ML models to interpret panoramic X-rays have been developed  specifically for individual tasks such as quadrant segmentation~\cite{Zhao2020,pati2021gandlf}, tooth detection~\cite{Chung2021}, dental enumeration~\cite{Lin2021,Tuzoff2019}, diagnosis of some abnormalities~\cite{Zhu2022,Krois2019}, as well as treatment planning~\cite{Yuksel2021}. Although many of these studies have achieved good results, three main issues still remain. \emph{(1) Multi-label detection:} there has not been an end-to-end model developed that gives all the necessary information for treatment planning by detecting abnormal teeth with dental enumeration and multiple diagnoses simultaneously~\cite{AbuSalim2022}. \emph{(2) Data availability:} to train a model that performs this task with high accuracy, a large set of fully annotated data is needed~\cite{Kumar2021}. Because labeling every tooth with all required classes may require expertise and take a long time, such kind of fully labeled large datasets do not always exist~\cite{Willemink2020}. For instance, we structure three different available annotated data hierarchically shown in Fig.~\ref{fig2}, using the Fédération Dentaire Internationale~(FDI) system. The first data is partially labeled because it only included quadrant information. The second data is also partially labeled but contains additional enumeration information along with the quadrant. The third data is fully labeled because it includes all quadrant-enumeration-diagnosis information for each abnormal tooth. Thus, conventional object detection algorithms would not be well applicable to this kind of hierarchically and partially annotated data~\cite{Shin2020}. \emph{(3) Model performance:} to the best of our knowledge, models designed to detect multiple diagnoses on panoramic X-rays have not achieved the same high level of accuracy as those specifically designed for individual tasks, such as tooth detection, dental enumeration, or detecting single abnormalities~\cite{Panetta2022}.

To circumvent the limitations of the existing methods, we propose a novel diffusion-based hierarchical multi-label object detection method to point out each abnormal tooth with dental enumeration and associated diagnosis concurrently on panoramic X-rays, see Fig.~\ref{fig1}. Due to the partial annotated and hierarchical characteristics of our data, we adapt a diffusion-based method \cite{chen2022diffusiondet} that formulates object detection as a denoising diffusion process from noisy boxes to object boxes. Compared to the previous object detection methods that utilize conventional weight transfer~\cite{bu2021gaia} or cropping strategies~\cite{shin2020hierarchical} for hierarchical learning,  the denoising process enables us to propose a novel hierarchical diffusion network by utilizing the inference from the previously trained model in hierarchical order to manipulate the noisy bounding boxes as in Fig.~\ref{fig1}. Besides, instead of pseudo labeling techniques~\cite{zhao2020object} for partially annotated data,  we develop a multi-label object detection method to learn efficiently from partial annotations and to give all the needed information about each abnormal tooth for treatment planning. Finally, we demonstrate the effectiveness of our multi-label detection method on partially annotated data and the efficacy of our proposed bounding box manipulation technique in diffusion networks for hierarchical data.% in Tab.~\ref{table1}. 

The contributions of our work are three-fold. (1) We propose a multi-label detector to learn efficiently from partial annotations and to detect the abnormal tooth with all three necessary classes, as shown in Fig~\ref{fig4} for treatment planning. (2) We rely on the denoising process of diffusion models~\cite{chen2022diffusiondet} and frame the detection problem as a hierarchical learning task by proposing a novel bounding box manipulation technique that outperforms conventional weight transfer as shown in Fig.~\ref{fig5}. (3) Experimental results show that our model with bounding box manipulation and multi-label detection significantly outperforms state-of-the-art object detection methods on panoramic X-ray analysis, as shown in Tab.~\ref{table1}. 

We have designed our approach to serve as a foundational baseline for the Dental Enumeration and Diagnosis on Panoramic X-rays Challenge (DENTEX), set to take place at MICCAI 2023. Remarkably, the data set and annotations we utilized for our method mirror exactly those employed for DENTEX~\cite{hamamci2023dentex}.

\section{Methods}
\label{section:methods}

\begin{figure}[htp]
    \begin{adjustbox}{height=0.9\textheight}
        \includegraphics{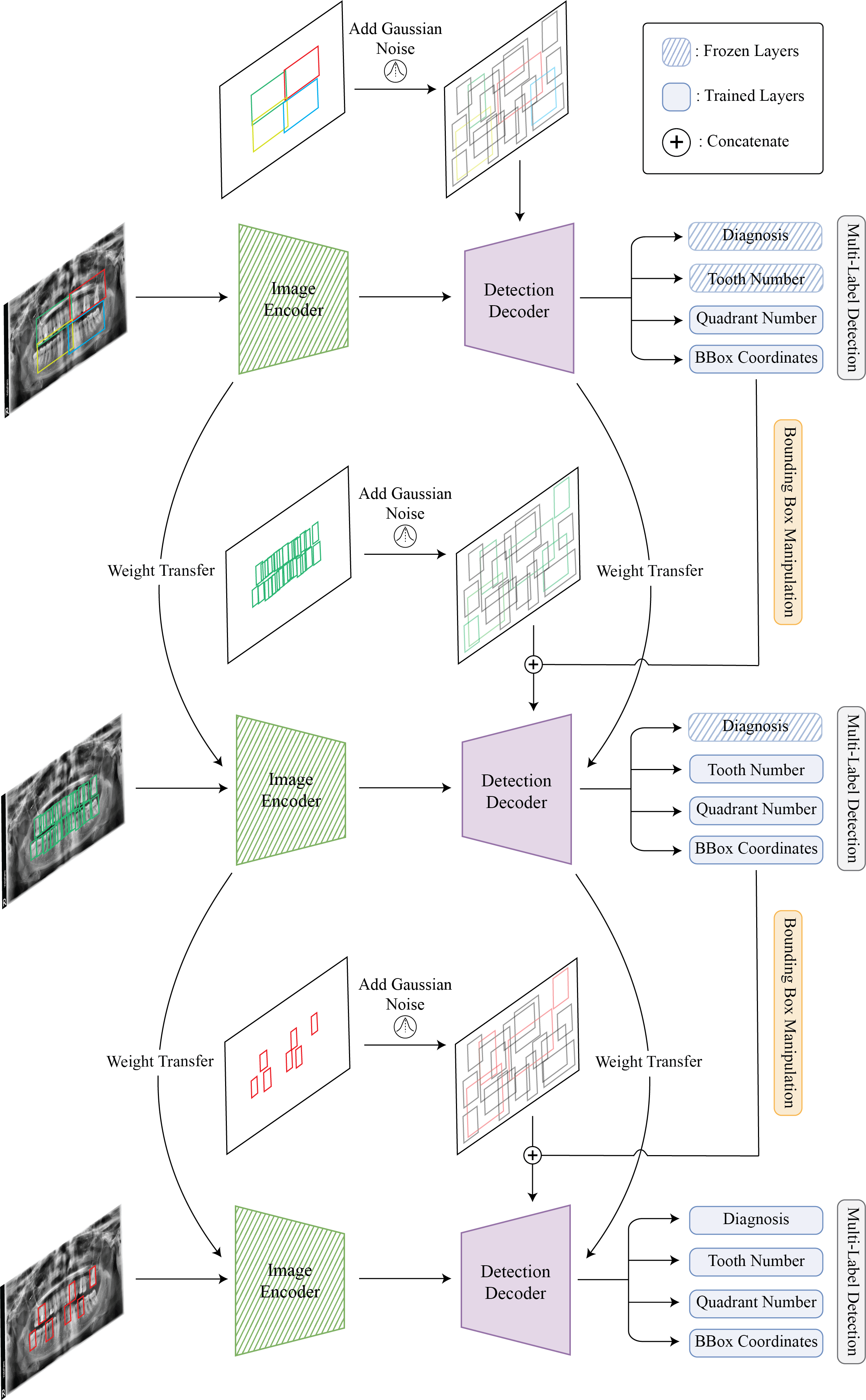}
    \end{adjustbox}
    \centering
    \caption{Our method relies on a hierarchical learning approach utilizing a combination of multi-label detection, bounding box manipulation, and weight transfer.} \label{fig1}
\end{figure}

Figure~\ref{fig1} illustrates our proposed framework. We utilize the DiffusionDet~\cite{chen2022diffusiondet} model, which formulates object detection as a denoising diffusion process from noisy boxes to object boxes. Unlike other state-of-the-art detection models, the denoising property of the model enables us to propose a novel manipulation technique to utilize a hierarchical learning architecture by using previously inferred boxes. Besides, to learn efficiently from partial annotations, we design a multi-label detector with adaptable classification layers based on available labels. 
%In addition, we implemented a weight transfer workflow that allows for incremental training, beginning with only the quadrant class and gradually expanding to include the enumeration and diagnosis classes.

%We propose DiffusionDet, a new framework that formulates object detection as a denoising diffusion process from noisy boxes to object boxes. During the training stage, object boxes diffuse from ground-truth boxes to random distribution, and the model learns to reverse this noising process. In inference, the model refines a set of randomly generated boxes to the output results in a progressive way.
%The primary contributions of our work are two-fold. 

\clearpage
\subsection{Base Model}

Our method employs the DiffusionDet~\cite{chen2022diffusiondet} that comprises two essential components, an image encoder that extracts high-level features from the raw image and a detection decoder that refines the box predictions from the noisy boxes using those features. The set of initial noisy bounding boxes is defined as: 
\begin{equation} \label{eqn:eq1}
 q(z_t|z_0) = \mathcal{N}(z_t|\sqrt{\bar\alpha_t}z_0,(1-\bar\alpha_t)I)\end{equation} 
where \begin{math} z_0\end{math} represents the input bounding box \begin{math}b\end{math}, and \begin{math}    b \in{ \mathbb{R}^{N\times4}}\end{math} is a set of bounding boxes, \begin{math}z_t\end{math} represents the latent noisy boxes, and \begin{math}\bar\alpha_t\end{math} represents the noise variance schedule. The DiffusionDet model~\cite{chen2022diffusiondet} \begin{math}f_\theta(z_t, t, x)\end{math}, is trained to predict the final bounding boxes defined as \begin{math}b^i = (c^i_x, c^i_y, w^i, h^i)\end{math} where \begin{math}(c^i_x, c^i_y)\end{math} are the center coordinates of the bounding box and \begin{math}(w^i, h^i)\end{math} are the width and height of the bounding boxes and category labels defined as \begin{math}y^i\end{math} for objects.

\subsection{Proposed Framework}
To improve computational efficiency during the denoising process, DiffusionDet~\cite{chen2022diffusiondet} is divided into two parts: an image encoder and a detection decoder. Iterative denoising is applied only for the detection decoder, using the outputs of the image encoder as a condition. Our method employs this approach with several adjustments, including multi-label detection and bounding box manipulation. Finally, we utilize conventional transfer learning for comparison.

\noindent
\textbf{Image Encoder.} Our method utilizes a Swin-transformer~\cite{swin_transformer} backbone pre-trained on the ImageNet-22k~\cite{imagenet} with a Feature Pyramid Network (FPN) architecture~\cite{FPN}  as it was shown to outperform convolutional neural network-based models such as ResNet50~\cite{resnet50}. We also apply pre-training to the image encoder using our unlabeled data, as it is not trained during the training process. We utilize SimMIM~\cite{SIMMIM} that uses masked image modeling to finetune the encoder.

\noindent
\textbf{Detection Decoder.} Our method employs a detection decoder that inputs noisy initial boxes to extract Region of Interest~(RoI) features from the encoder-generated feature map and predicts box coordinates and classifications using a detection head. However, our detection decoder has several differences from DiffusionDet~\cite{chen2022diffusiondet}. Our proposed detection decoder (1) has three classification heads instead of one, which allows us to train the same model with partially annotated data by freezing the heads according to the unlabeled classes, (2) employs manipulated bounding boxes to extract RoI features, and (3) leverages transfer learning from previous training steps.

\begin{figure}[ht]
    \includegraphics[scale=.225]{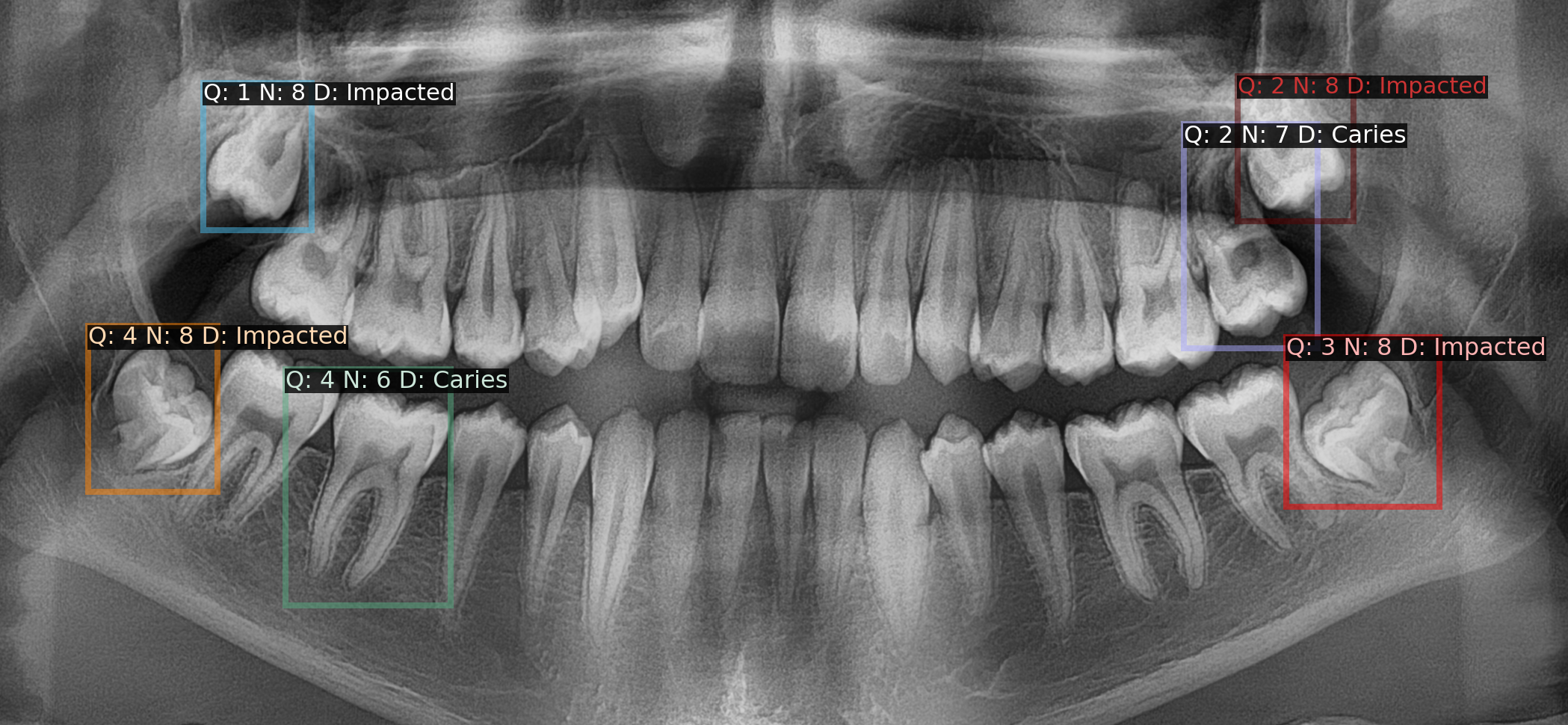}
    \centering
    \caption{Output from our final model showing well-defined boxes for diseased teeth with corresponding quadrant (Q), enumeration (N), and diagnosis (D) labels.} \label{fig4}
    \vspace{-0.00cm}
\end{figure}

\noindent
\textbf{Multi-Label Detection.} We utilize three classification heads as quadrant-enumeration-diagnosis for each bounding box and freeze the heads for the unlabeled classes, shown in Fig.~\ref{fig1}. Our model denoted by $f_\theta$ is trained to predict:

\begin{equation} \label{eqn:eq2}
    f_{\theta}(z_t,t,x,h_q,h_e,h_d)=\left\{\begin{matrix}
 (y_q^i ,b^i),& & h_q=1, h_e=0,h_d=0 & & & & (a)&\\
 (y_q^i ,y_e^i,b^i),& & h_q=1, h_e=1,h_d=0 & & & & (b)& \\ 
 (y_q^i ,y_e^i ,y_d^i,b^i),& & h_q=1, h_e=1,h_d=1 & & & & (c)&
 
\end{matrix}\right.
\end{equation}
where $y_q^i$, $y_e^i$, and $y_d^i$ represent the bounding box classifications for quadrant, enumeration, and diagnosis, respectively, and $h_q$, $h_e$, and $h_d$  represent binary indicators of whether the labels are present in the training dataset. By adapting this approach, we leverage the full range of available information and improve our ability to handle partially labeled data. This stands in contrast to conventional object detection methods, which rely on a single classification head for each bounding box~\cite{wu2019detectron2} and may not capture the full complexity of the underlying data. Besides, this approach enables the model to detect abnormal teeth with all three necessary classes for clinicians to plan the treatment, as seen in Fig.~\ref{fig4}. 

% \textbf{Transfer Learning:} We employed a transfer learning strategy to maximize partially labeled data during training. Our approach starts with training the model onxcv quadrant labels, transferring the learned weights to enumeration training, and then using those weights for the final diagnosis trainin g. This hierarchical approach, where each step builds on the previous, makes transfer learning a critical component of our pipeline.
\noindent
\textbf{Bounding Box Manipulation.} Instead of completely noisy boxes, we use manipulated bounding boxes to extract RoI features from the encoder-generated feature map and to learn efficiently from hierarchical annotations as shown in Fig.~\ref{fig1}. Specifically, to train the model (b) in Eq.~(\ref{eqn:eq2}), we concatenate the noisy boxes described in Eq.~(\ref{eqn:eq1}) with the boxes inferred from the model (a) in Eq.~(\ref{eqn:eq2}) with a score greater than 0.5. Similarly, we manipulate the denoising process during the training of the model (c) in Eq.~(\ref{eqn:eq2}) by concatenating the noisy boxes with boxes inferred from the model (b) in Eq.~(\ref{eqn:eq2}) with a score greater than 0.5. The set of manipulated boxes $b_m$, and \begin{math}    b_m \in{ \mathbb{R}^{N\times4}}\end{math}, can be defined as \begin{math}
b_{m} = [b_{n}[:-k], b_{i}]
\end{math}, where $b_n$, and \begin{math}    b_n \in{ \mathbb{R}^{N\times4}}\end{math}, represents the set of noisy boxes and, $b_i$, and \begin{math}    b_i \in{ \mathbb{R}^{k\times4}}\end{math}, represents the set of inferred boxes from the previous training. Our framework utilizes completely noisy boxes during the inference.

\clearpage
\section{Experiments and Results}
\label{section:results}
We evaluate models' performances using a combination of Average Recall (AR) and Average Precision (AP) scores with various Intersection over Union (IoU) thresholds. This included AP\textsubscript{[0.5,0.95]}, AP\textsubscript{50}, AP\textsubscript{75}, and separate AP scores for large objects (AP\textsubscript{l}), and medium objects (AP\textsubscript{m}). 
%AR was calculated by doubling the area under the Recall x IoU curve for IoU ranging from 0.5 to 1.0. AP was calculated by computing the area under the Precision x Recall curve using the AUC method.

\begin{table}[ht]
\centering
\begin{adjustbox}{width=0.87\textwidth}

\begin{tabular}{@{}llllllll@{}}
\toprule
\rowcolor[HTML]{FFFFFF} 
Method                                                     & {\color[HTML]{FFFFFF} .....} & AR & AP &AP\textsubscript{50} & AP\textsubscript{75} & AP\textsubscript{m} & AP\textsubscript{l} \\ \midrule
\rowcolor[HTML]{FFFFFF} 
\multicolumn{8}{c}{\cellcolor[HTML]{FFFFFF}Quadrant}                                                                            \\
\rowcolor[HTML]{FFFFFF} 
RetinaNet\cite{retinanet}                            & \cellcolor[HTML]{FFFFFF}     & 0.604  & 25.1  & 41.7    & 28.8    & 32.9   & 25.1      \\
\rowcolor[HTML]{FFFFFF} 
Faster R-CNN\cite{Ren2017}                         & \cellcolor[HTML]{FFFFFF}     & 0.588  & 29.5  & 48.6    & 33.0    & 39.9   & 29.5     \\
\rowcolor[HTML]{FFFFFF} 
DETR\cite{10.1007/978-3-030-58452-8_13}                                 & \cellcolor[HTML]{FFFFFF}     & 0.659  & 39.1  & 60.5    & 47.6    & 55.0   & 39.1     \\

\rowcolor[HTML]{FFFFFF} 
Base~(DiffusionDet)\cite{chen2022diffusiondet}                                         &                              & 0.677  & 38.8  & 60.7    & 46.1    & 39.1   & 39.0     \\
\rowcolor[HTML]{EFEFEF} 
Ours w/o Transfer                                          &                              & 0.699  & 42.7  & 64.7    & \textbf{52.4}    & 50.5   & 42.8      \\
\rowcolor[HTML]{EFEFEF} 
Ours w/o Manipulation                                      &                              & \textbf{0.727}  & 40.0  & 60.7    & 48.2    & 59.3   & 40.0      \\
\rowcolor[HTML]{EFEFEF} 
Ours w/o Manipulation and Transfer                         &                              & 0.658  & 38.1  & 60.1    & 45.3    & 45.1   & 38.1      \\
\rowcolor[HTML]{EFEFEF} 
Ours (Manipulation+Transfer+Multilabel)                    &                               & 0.717  & \textbf{43.2}    & \textbf{65.1}    & 51.0   & \textbf{68.3}   & \textbf{43.1}     \\ \midrule
\rowcolor[HTML]{FFFFFF} 
\multicolumn{8}{c}{\cellcolor[HTML]{FFFFFF}Enumeration}                                                                             \\

\rowcolor[HTML]{FFFFFF} 
RetinaNet\cite{retinanet}                            & \cellcolor[HTML]{FFFFFF}     & 0.560  & 25.4  & 41.5    & 28.5    & 55.1   & 25.2      \\
\rowcolor[HTML]{FFFFFF} 
Faster R-CNN\cite{Ren2017}                         & \cellcolor[HTML]{FFFFFF}     & 0.496  & 25.6  & 43.7    & 27.0    & 53.3   & 25.2     \\
\rowcolor[HTML]{FFFFFF} 
DETR\cite{10.1007/978-3-030-58452-8_13}                                 & \cellcolor[HTML]{FFFFFF}     & 0.440  & 23.1  & 37.3    & 26.6    & 43.4   & 23.0      \\

\rowcolor[HTML]{FFFFFF} 
Base~(DiffusionDet)\cite{chen2022diffusiondet}                                         & \cellcolor[HTML]{FFFFFF}     & 0.617  & 29.9  & 47.4    & 34.2    & 48.6   & 29.7     \\
\rowcolor[HTML]{EFEFEF} 
\cellcolor[HTML]{EFEFEF}Ours w/o Transfer                  &                              & 0.648  & \textbf{32.8}  & \textbf{49.4}    & \textbf{39.4}    & \textbf{60.1}  & \textbf{32.9}      \\
\rowcolor[HTML]{EFEFEF} 
\cellcolor[HTML]{EFEFEF}Ours w/o Manipulation  &                              & 0.662  & 30.4  & 46.5    & 36.6    & 58.4   & 30.5      \\

\rowcolor[HTML]{EFEFEF} 
\cellcolor[HTML]{EFEFEF}Ours w/o Manipulation and Transfer             &                              & 0.557  & 26.8  & 42.4    & 29.5    & 51.4   & 26.5      \\

\rowcolor[HTML]{EFEFEF} 
Ours (Manipulation+Transfer+Multilabel)                    & \cellcolor[HTML]{EFEFEF}     & \textbf{0.668}  & 30.5  & 47.6    & 37.1    & 51.8   & 30.4      \\
\midrule
\rowcolor[HTML]{FFFFFF} 
\multicolumn{8}{c}{\cellcolor[HTML]{FFFFFF}Diagnosis}                                                                               \\

\rowcolor[HTML]{FFFFFF} 
RetinaNet\cite{retinanet}                            & \cellcolor[HTML]{FFFFFF}     & 0.587  & 32.5  & 54.2    & 35.6   & 41.7   & 32.5      \\
\rowcolor[HTML]{FFFFFF} 
Faster R-CNN\cite{Ren2017}                         & \cellcolor[HTML]{FFFFFF}     & 0.533  & 33.2  & 54.3    & 38.0   & 24.2   & 33.3      \\
\rowcolor[HTML]{FFFFFF} 
DETR\cite{10.1007/978-3-030-58452-8_13}                                 & \cellcolor[HTML]{FFFFFF}     & 0.514  & 33.4  & 52.8    & 41.7    & 48.3   & 33.4      \\

\rowcolor[HTML]{FFFFFF} 
Base~(DiffusionDet)\cite{chen2022diffusiondet}                                         & \cellcolor[HTML]{FFFFFF}     & 0.644  & 37.0  & 58.1    & 42.6    & 31.8   & 37.2      \\
\rowcolor[HTML]{EFEFEF} 
\cellcolor[HTML]{EFEFEF}Ours w/o Transfer                  &                              & 0.669  & \textbf{39.4}  & \textbf{61.3}    & \textbf{47.9}    & \textbf{49.7}   & \textbf{39.5}      \\
\rowcolor[HTML]{EFEFEF} 
\cellcolor[HTML]{EFEFEF}Ours w/o Manipulation              &                              & 0.688  & 36.3  & 55.5    & 43.1    & 45.6   & 37.4      \\
\rowcolor[HTML]{EFEFEF} 
\cellcolor[HTML]{EFEFEF}Ours w/o Manipulation and Transfer &                              & 0.648  & 37.3  & 59.5    & 42.8    & 33.6   & 36.4      \\
\rowcolor[HTML]{EFEFEF} 
Ours (Manipulation+Transfer+Multilabel)                    & \cellcolor[HTML]{EFEFEF}     & \textbf{0.691}  & 37.6  & 60.2    & 44.0    & 36.0   & 37.7      \\ \bottomrule
\end{tabular}

\end{adjustbox}
\vspace{0.3cm}
\caption{\label{table1}
% Table showing the object detection results for our approach and previous state-of-the-art methods. 
Our method outperforms state-of-the-art methods, and our bounding box manipulation approach outperforms the weight transfer. Results shown here indicate the different tasks in the test set which is multi-labeled (quadrant-enumeration-diagnosis) for abnormal tooth detection.}
\vspace{-0.6cm}
\end{table}
\noindent
\textbf{Data.} All panoramic X-rays were acquired from patients above 12 years of age using the VistaPano S X-ray unit (Durr Dental, Germany). To ensure patient privacy and confidentiality, panoramic X-rays were randomly selected from the hospital's database without considering any personal information. 

To effectively utilize FDI system~\cite{InternationalDentalFederation2009},  three distinct types of data are organized hierarchically as in Fig.~\ref{fig2} (a) 693 X-rays labeled only for quadrant detection, (b) 634 X-rays labeled for tooth detection with both quadrant and tooth enumeration classifications, and (c) 1005 X-rays fully labeled for diseased tooth detection with quadrant, tooth enumeration, and diagnosis classifications. In the diagnosis, there are four specific classes corresponding to four different diagnoses: caries, deep caries, periapical lesions, and impacted teeth. The remaining 1571 unlabeled X-rays are used for pre-training. All necessary permissions were obtained from the ethics committee.

\begin{figure}[ht]

\begin{minipage}{.31\linewidth}
    \centering
    Quadrant Metrics
\end{minipage}%
\begin{minipage}{.025\linewidth}
    \centering
    \phantom{\includegraphics[width=\linewidth]{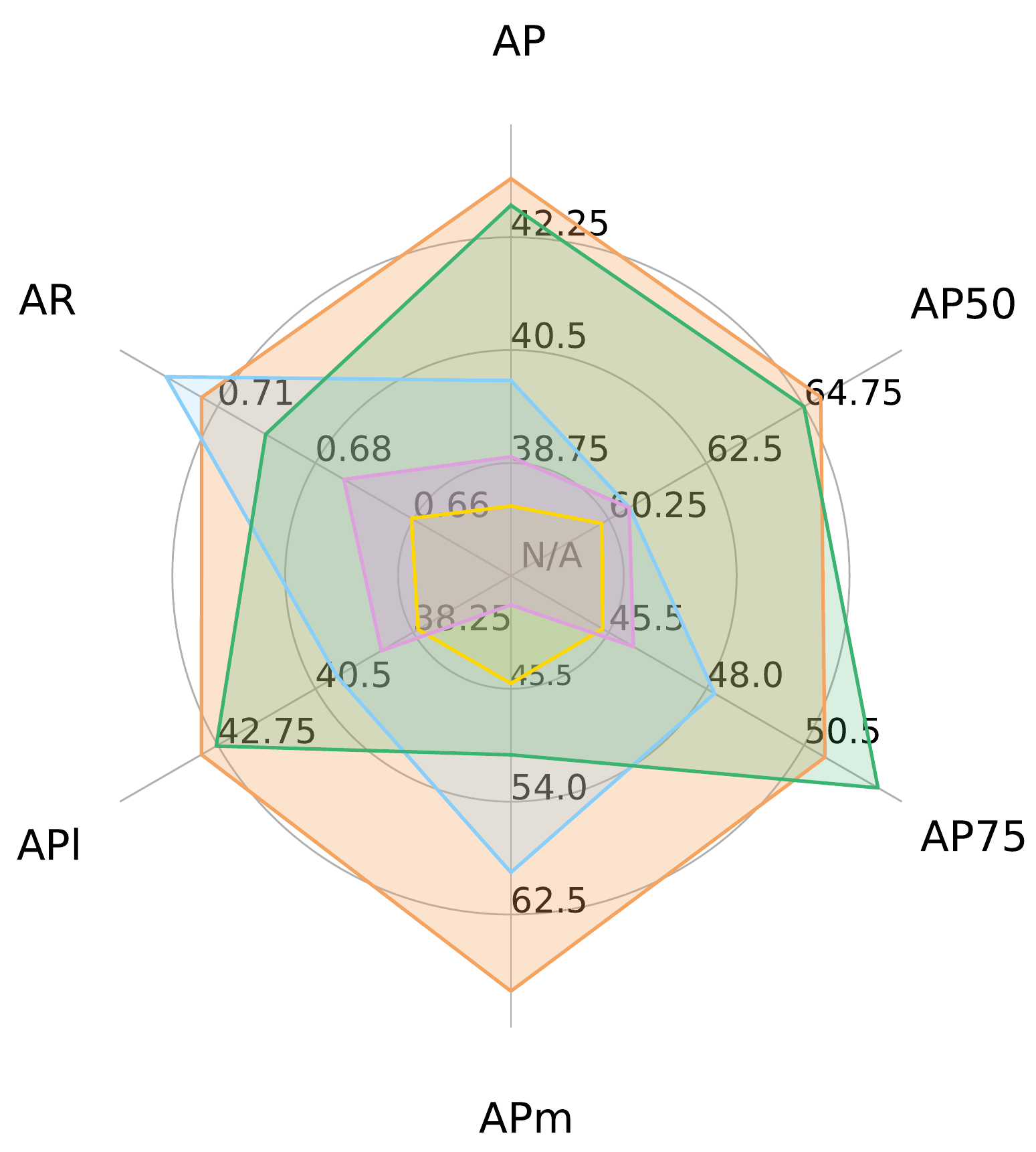}}
\end{minipage}%
\begin{minipage}{.31\linewidth}
    \centering
    Enumeration Metrics
\end{minipage}%
\begin{minipage}{.025\linewidth}
    \centering
    \phantom{\includegraphics[width=\linewidth]{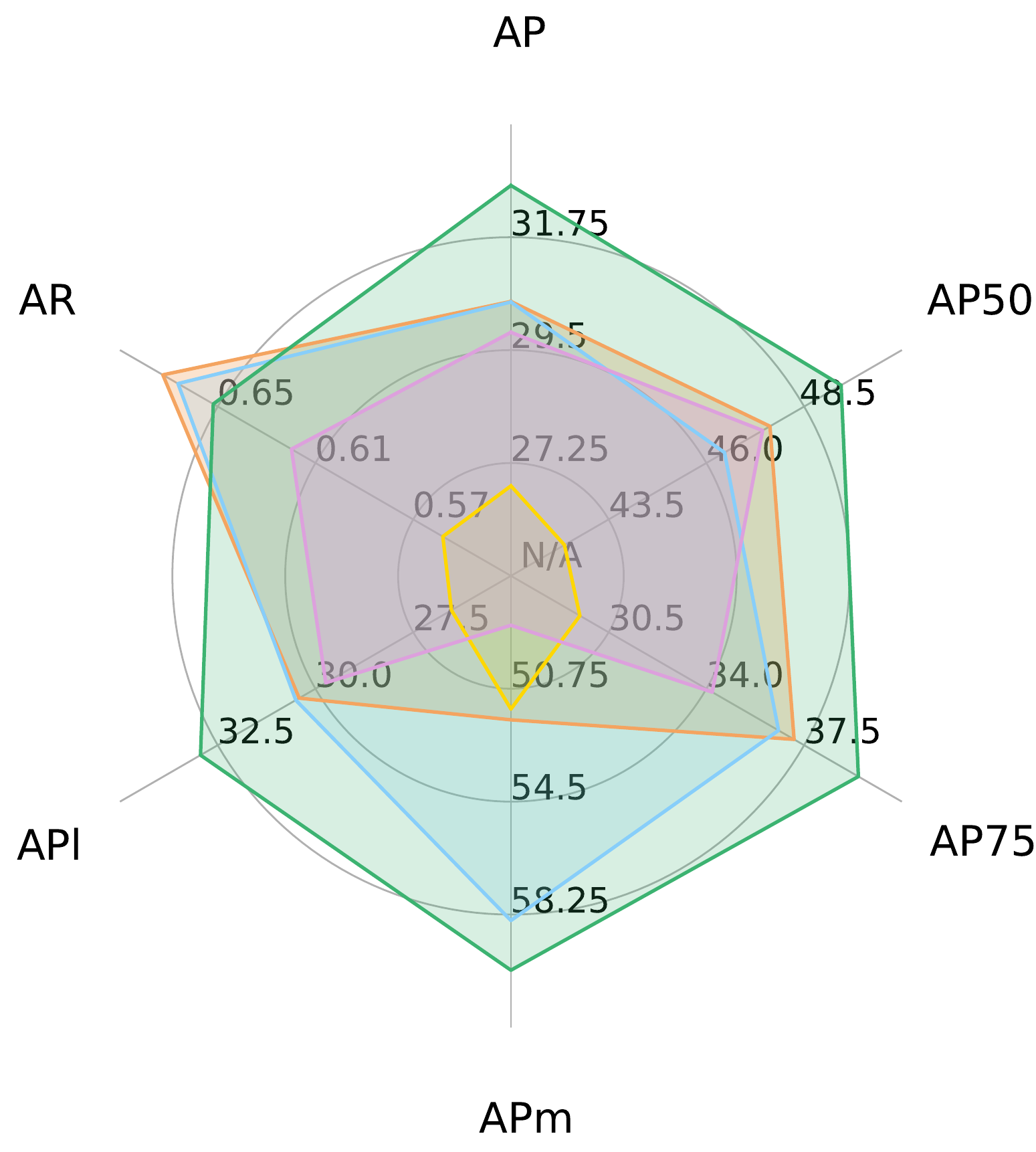}}
\end{minipage}%
\begin{minipage}{.31\linewidth}
    \centering
    Diagnosis Metrics
\end{minipage}%
% \begin{minipage}{.1\linewidth}
%     \centering
%     \phantom{\includegraphics[width=\linewidth]{figures/fig5b.pdf}}
% \end{minipage}%

\begin{minipage}{.31\linewidth}
    \centering
    \includegraphics[width=\linewidth]{figures/fig5a.pdf}
\end{minipage}%
\begin{minipage}{.025\linewidth}
    \centering
    \phantom{\includegraphics[width=\linewidth]{figures/fig5a.pdf}}
\end{minipage}%
\begin{minipage}{.31\linewidth}
    \centering
    \includegraphics[width=\linewidth]{figures/fig5b.pdf}
\end{minipage}%
\begin{minipage}{.025\linewidth}
    \centering
    \phantom{\includegraphics[width=\linewidth]{figures/fig5b.pdf}}
\end{minipage}%
\begin{minipage}{.31\linewidth}
    \centering
    \includegraphics[width=\linewidth]{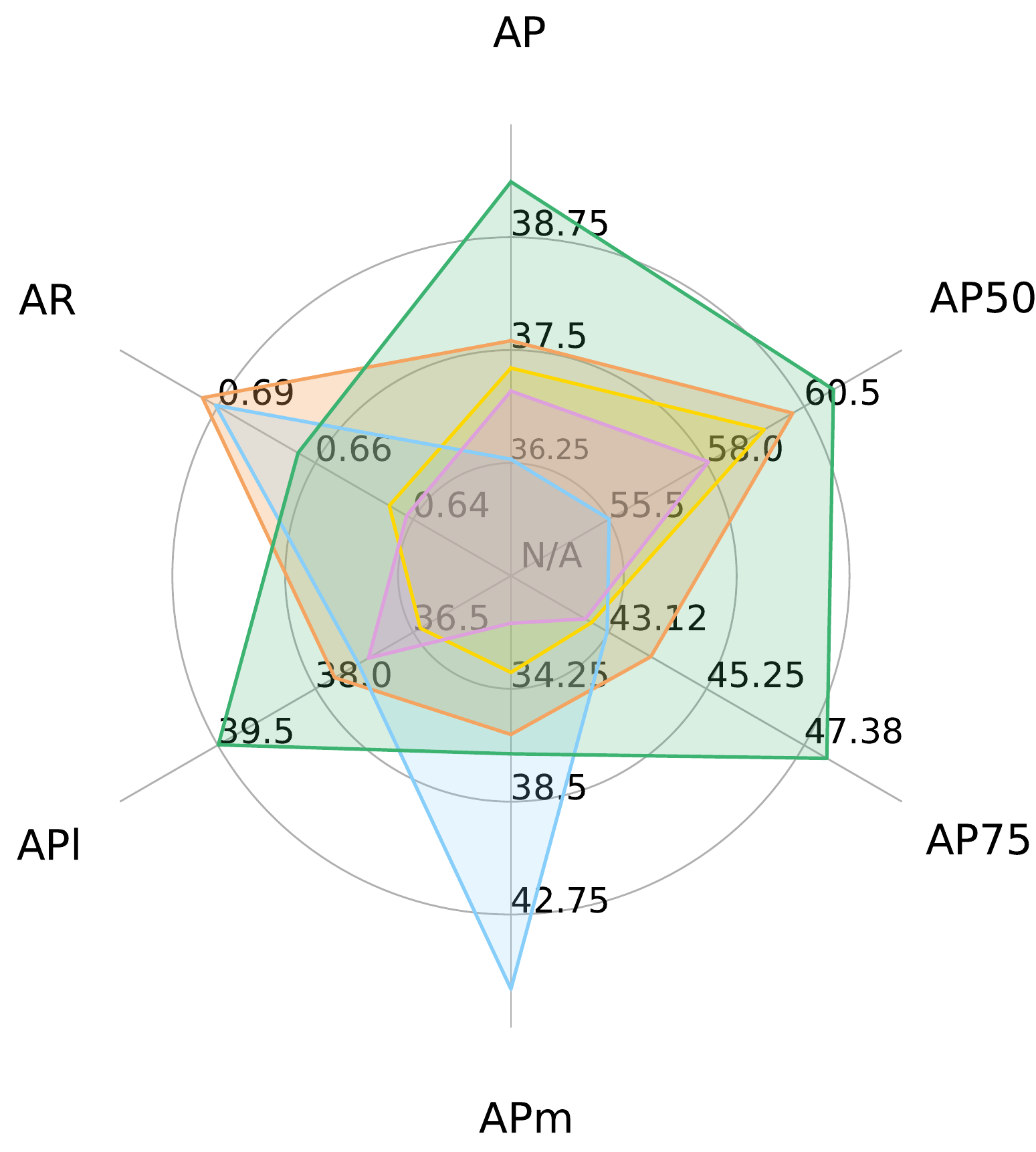}
\end{minipage}%\par\medskip
\begin{minipage}{1\linewidth}
    \vspace{3.5cm}
    \hspace{-5.7cm}
    \includegraphics[width=0.28\linewidth]{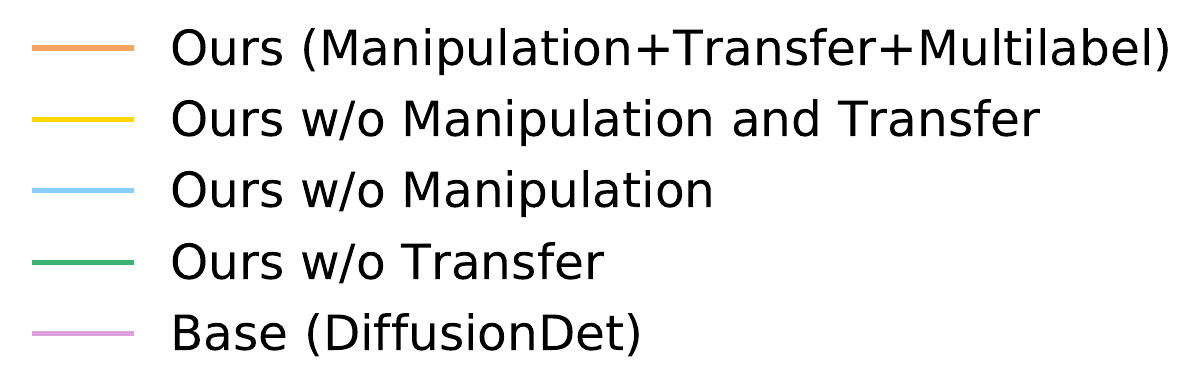}
\end{minipage}%
\caption{The results of the ablation study reveals that our bounding box manipulation method outperforms conventional weight transfer.}
\label{fig5}
\end{figure}

\noindent
\textbf{Experimental Design.} To evaluate our proposed method, we conduct two experiments: (1) Comparison with state-of-the-art object detection models, including DETR~\cite{10.1007/978-3-030-58452-8_13}, Faster R-CNN \cite{Ren2017}, RetinaNet~\cite{retinanet}, and DiffusionDet~\cite{chen2022diffusiondet} in Tab.~\ref{table1}. (2) A comprehensive ablation study to assess the effect of our modifications to DiffusionDet in hierarchical detection performance in Fig.~\ref{fig5}.

\noindent
\textbf{Evaluation.} Figure~\ref{fig4} presents the output prediction of the final trained model. As depicted in the figure, the model effectively assigns three distinct classes to each well-defined bounding box. Our approach that utilizes novel box manipulation and multi-label detection, significantly outperforms state-of-the-art methods. The box manipulation approach specifically leads to significantly higher AP and AR scores compared to other state-of-the-art methods, including RetinaNet, Faster-R-CNN, DETR, and DiffusionDet. Although the impact of conventional transfer learning on these scores can vary depending on the data, our bounding box manipulation outperforms it. Specifically, the bounding box manipulation approach is the sole factor that improves the accuracy of the model, while weight transfer does not improve the overall accuracy, as shown in Fig.~\ref{fig5}. %Further examples of single and double-labeled detection outputs that are used in the hierarchical training can be found in the Supplementary. 

\vspace{-0.5cm}
\noindent
\textbf{Ablation Study.}
Our ablation study results, shown in Fig.~\ref{fig5} and Tab.~\ref{table1}, indicate that our approaches have a synergistic impact on the detection model's accuracy, with the highest increase seen through bounding box manipulation. We systematically remove every combination of bounding box manipulation and weight transfer, to demonstrate the efficacy of our methodology. Conventional transfer learning does not positively affect the models’ performances compared to the bounding box manipulation, especially for enumeration and diagnosis.

\section{Discussion and Conclusion}
\label{section:discussion}
In this paper, we introduce a novel diffusion-based multi-label object detection framework to overcome one of the significant obstacles to the clinical application of ML models for medical and dental diagnosis, which is the difficulty in getting a large volume of fully labeled data. Specifically, we propose a novel bounding box manipulation technique during the denoising process of the diffusion networks with the inference from the previously trained model to take advantage of hierarchical data. Moreover, we utilize a multi-label detector to learn efficiently from partial annotations and to assign all necessary classes to each box for treatment planning. Our framework outperforms state-of-the-art object detection models for training with hierarchical and partially annotated panoramic X-ray data. 

From the clinical perspective, we develop a novel framework that simultaneously points out abnormal teeth with dental enumeration and associated diagnosis on panoramic dental X-rays with the help of our novel diffusion-based hierarchical multi-label object detection method. With some limits due to partially annotated and limited amount of data, our model that provides three necessary classes for treatment planning has a wide range of applications in the real world, from being a clinical decision support system to being a guide for dentistry students.

%
% ---- Bibliography ----

% BibTeX users should specify bibliography style 'splncs04'.
% References will then be sorted and formatted in the correct style.

\newpage
\noindent
\begin{center}\large{\textbf{Supplementary Material of Diffusion-Based Hierarchical Multi-Label Object Detection to Analyze Panoramic Dental X-rays}}
\end{center}

% \enlargethispage{2cm}
\begin{figure}[!htp]
    \centering
    % \advance\leftskip-2.5cm
    \includegraphics[width=1\linewidth]{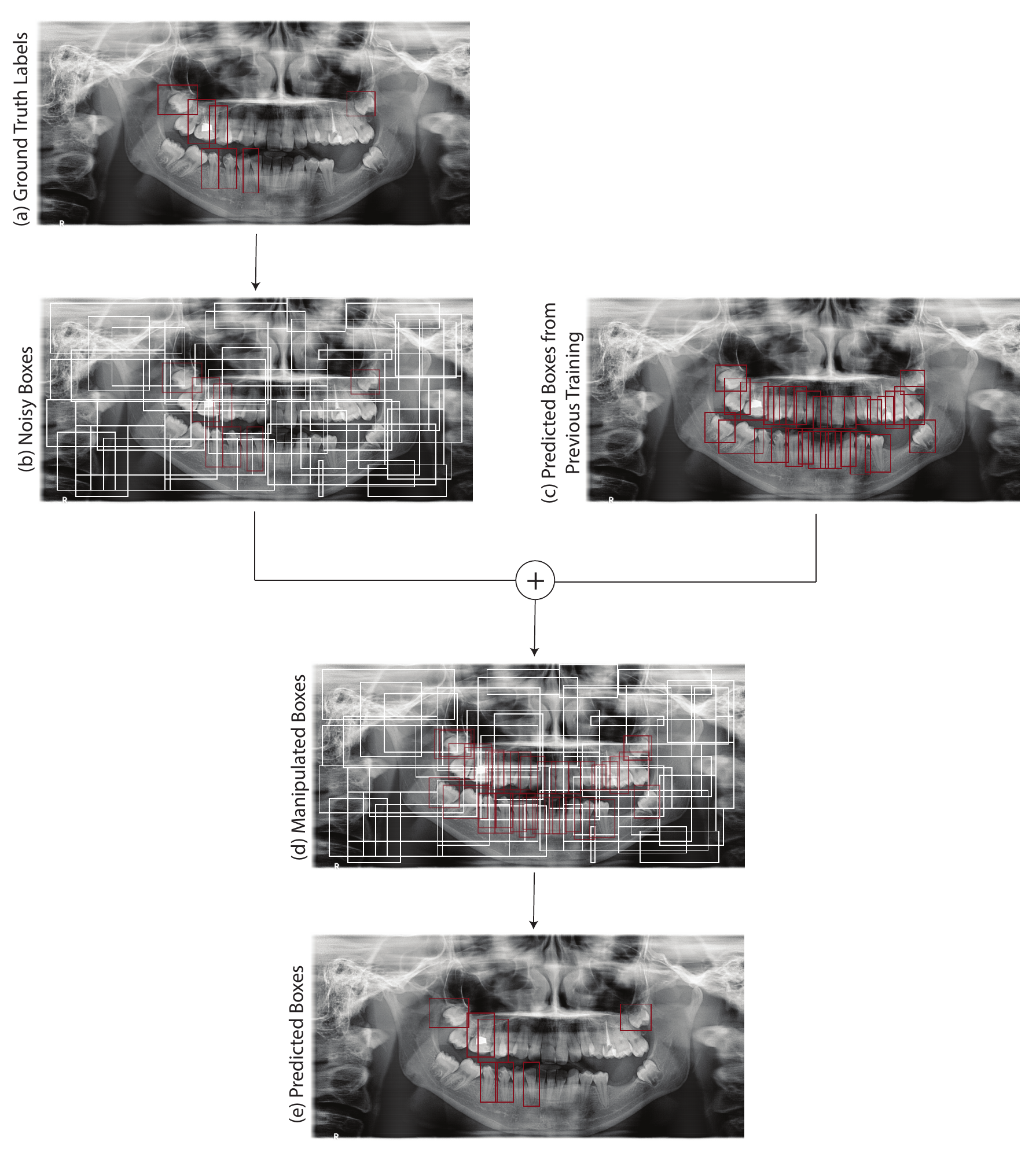}
    \caption{Figure showing the bounding box manipulation for the multi-label (quadrant-enumeration-diagnosis) abnormal tooth detection. Our bounding box manipulation method combines the boxes from the previously trained model for quadrant-enumeration with the noisy boxes. The process is very similar for the quadrant-enumeration in which quadrant boxes are used for the manipulation.}
    \label{supplementary_4}
\end{figure}
\clearpage

\begin{table}[]
\centering
\begin{tabular}{@{}llllllllllll@{}}
\toprule
 & Detection Model             &  &  & Image Encoder Backbone &  &  & Iterations   &  &  & Learning Rate &  \\ \midrule
\rowcolor[HTML]{EFEFEF} 
 & {\color[HTML]{000000} Ours} &  &  & FPN-Swin Transformer   &  &  & 40000        &  &  & 0.000025      &  \\
 & DiffusionDet                &  &  & FPN-Swin Transformer   &  &  & 40000        &  &  & 0.000025      &  \\
 & Faster R-CNN                &  &  & ResNet101              &  &  & 40000        &  &  & 0.02          &  \\
 & RetinaNet                   &  &  & ResNet101              &  &  & 40000        &  &  & 0.01          &  \\
 & DETR                        &  &  & ResNet50               &  &  & 300(epochs) &  &  & 0.0001        &  \\ \bottomrule
\end{tabular}
\vspace{0.3cm}
\caption{\label{table1s}
Different detection models are utilized for comparison with our method. The best test metrics for each model are selected for the results. All models are trained with randomly cropped and resized panoramic X-rays with a batch size of 16. All training is done on a single NVIDIA RTX A6000 48 GB GPU.}
\end{table}

\begin{table}[]
\centering
\begin{tabular}{@{}llllcllcllcl@{}}
\toprule
 & Dataset                         &  &  & Training &  &  & Validation &  &  & Testing &  \\ \midrule
\rowcolor[HTML]{FFFFFF} 
 & {\color[HTML]{000000} Quadrant} &  &  & 590      &  &  & 103        &  &  & N/A     &  \\
 & Quadrant-Enumeration            &  &  & 539      &  &  & 95         &  &  & N/A     &  \\
 & Quadrant-Enumeration-Diagnosis  &  &  & 705      &  &  & 50         &  &  & 250     &  \\ \bottomrule
\end{tabular}
\vspace{0.3cm}
\caption{\label{table1ss}
To ensure accurate testing of all models, we only use fully labeled data with quadrant-enumeration-diagnosis for abnormal tooth detection. We do not utilize quadrant or quadrant-enumeration data for testing. Our diagnosis labels have four specific classes: caries, deep caries, periapical lesions, and impacted.}
\end{table}

\begin{figure}[ht!]

\begin{minipage}{.31\linewidth}
    \centering
    (a) Quadrant detection with quadrant labels
\end{minipage}%
\begin{minipage}{.025\linewidth}
    \centering
    \phantom{\includegraphics[width=\linewidth]{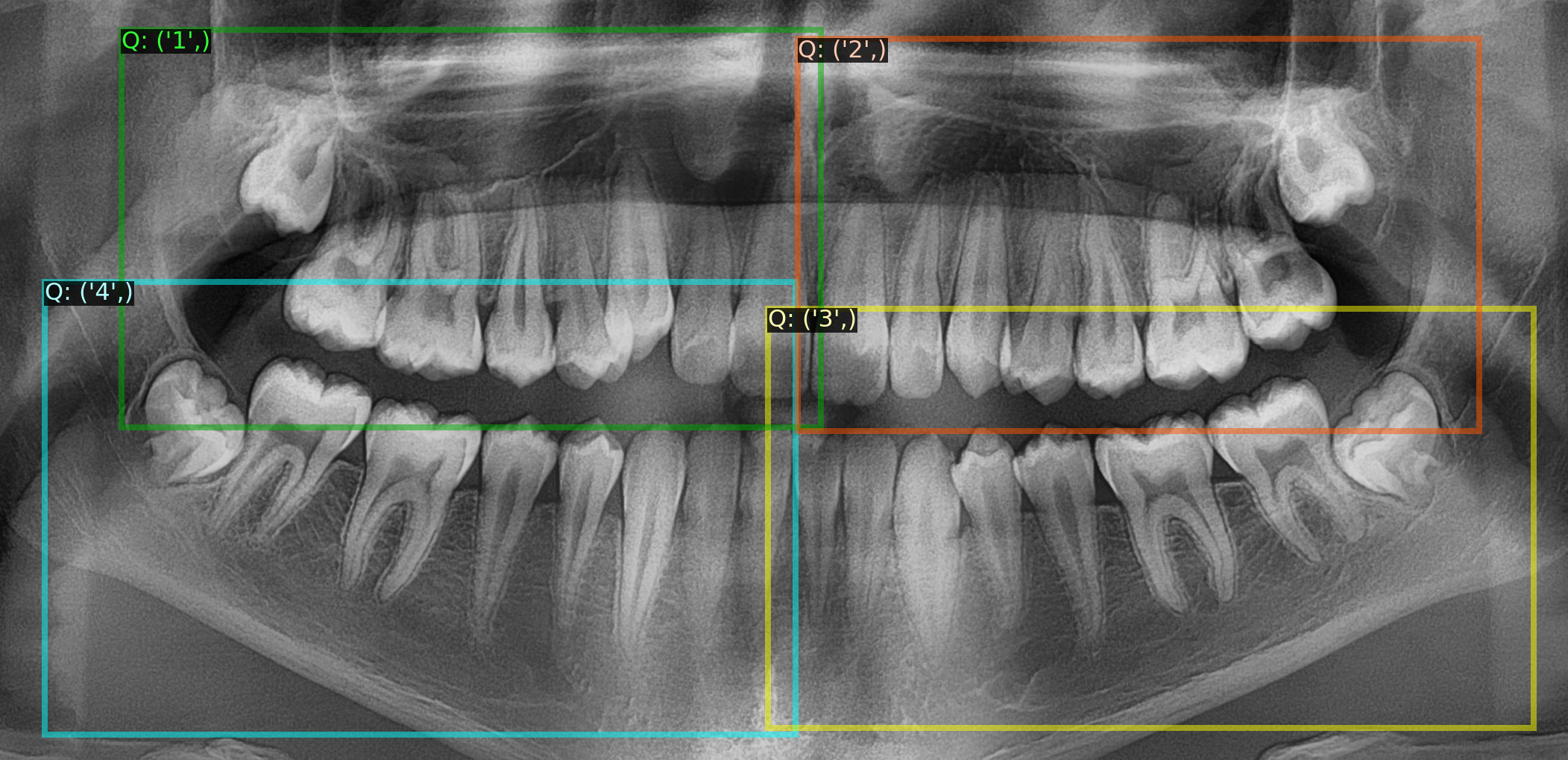}}
\end{minipage}%
\begin{minipage}{.31\linewidth}
    \centering
    (b) Tooth detection with quadrant-enumeration labels
\end{minipage}%
\begin{minipage}{.025\linewidth}
    \centering
    \phantom{\includegraphics[width=\linewidth]{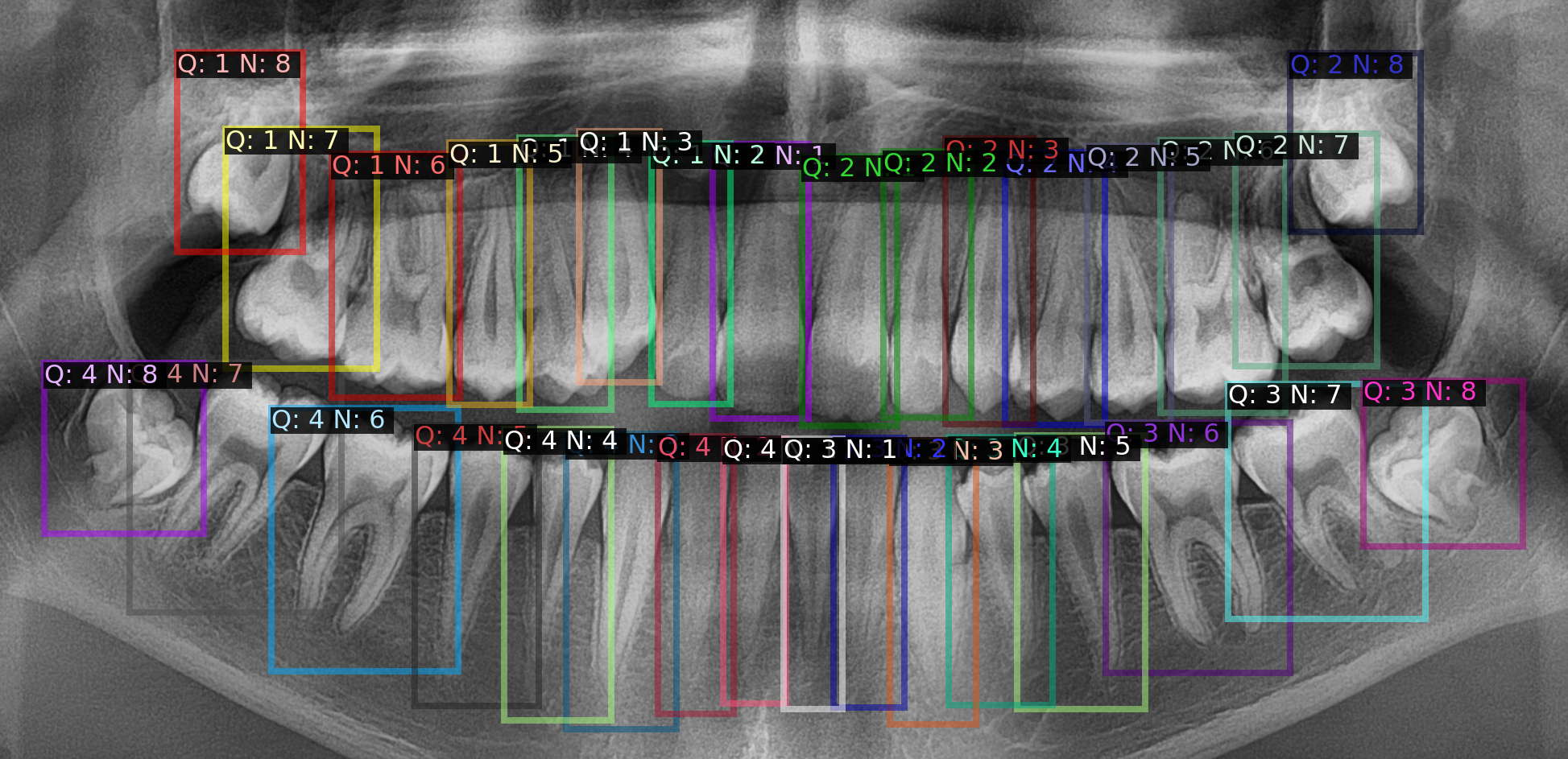}}
\end{minipage}%
\begin{minipage}{.31\linewidth}
    \centering
    (c) Abnormal tooth detection with quadrant-enumeration-diagnosis
\end{minipage}%

\begin{minipage}{.31\linewidth}
    \centering
    \includegraphics[width=\linewidth]{figures/supplementary_a.png}
\end{minipage}%
\begin{minipage}{.025\linewidth}
    \centering
    \phantom{\includegraphics[width=\linewidth]{figures/supplementary_b.png}}
\end{minipage}%
\begin{minipage}{.31\linewidth}
    \centering
    \includegraphics[width=\linewidth]{figures/supplementary_b.png}
\end{minipage}%
\begin{minipage}{.025\linewidth}
    \centering
    \phantom{\includegraphics[width=\linewidth]{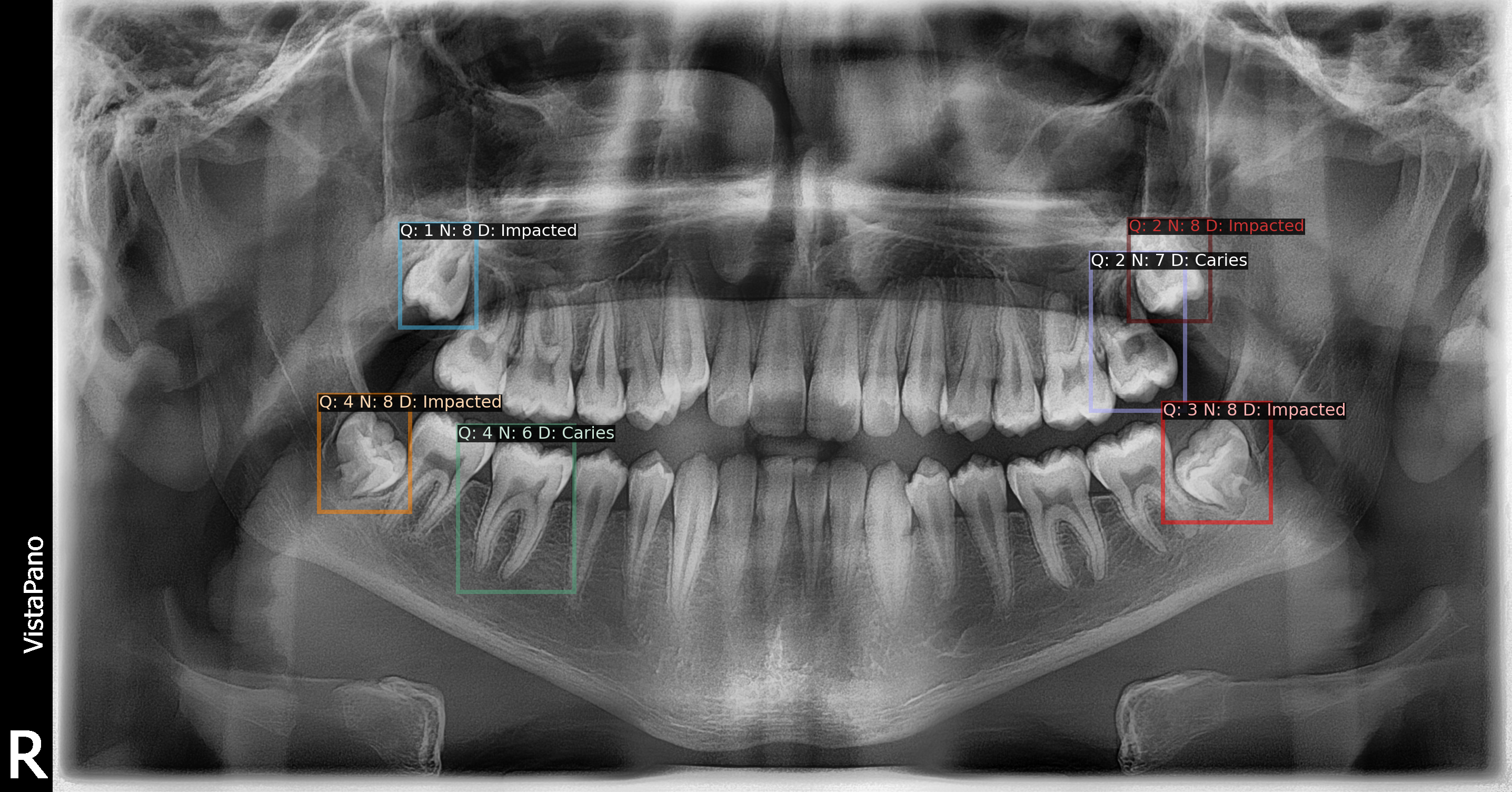}}
\end{minipage}%
\begin{minipage}{.31\linewidth}
    \centering
    \includegraphics[width=\linewidth]{figures/supplementary_c_zoomed.png}
\end{minipage}%\par\medskip
\caption{Example inferences during hierarchical training. (a) is used to manipulate noisy boxes during the training for (b). (b) is used to manipulate noisy boxes during the training for (c). (c) is the output of the final model.}
\label{supplementary_3}
\end{figure}

\begingroup
\bibliographystyle{splncs04}
\bibliography{paper}
\endgroup

\end{document}